\documentclass{article}

    \PassOptionsToPackage{numbers, compress}{natbib}
\usepackage[preprint]{neurips_2022}

\usepackage[utf8]{inputenc} % allow utf-8 input
\usepackage[T1]{fontenc}    % use 8-bit T1 fonts
\usepackage{hyperref}       % hyperlinks
\usepackage{url}            % simple URL typesetting
\usepackage{booktabs}       % professional-quality tables
\usepackage{amsfonts}       % blackboard math symbols
\usepackage{nicefrac}       % compact symbols for 1/2, etc.
\usepackage{microtype}      % microtypography
\usepackage{xcolor}         % colors
\usepackage{graphicx}
\usepackage{svg}
\usepackage{amsmath}
\usepackage{float}
\title{MetricGold: Leveraging Text-To-Image Latent Diffusion Models for Metric Depth}

\author{%
  Ansh Shah\\
  IIIT Hyderabad\\
  \And
  K. Madhava Krishna \\
  IIIT Hyderabad \\
}

\begin{document}

\newcommand{\projectpage}{\href{https://github.com/AnshShah3009/MetricGold}{Code: https://github.com/AnshShah3009/MetricGold}}

\maketitle
\vspace{-0.5cm}
\begin{center}
\projectpage
\end{center}

\begin{figure}[h]  % 'h' means "here", placing the figure at this location
    \centering
    \includegraphics[width=0.9\textwidth]{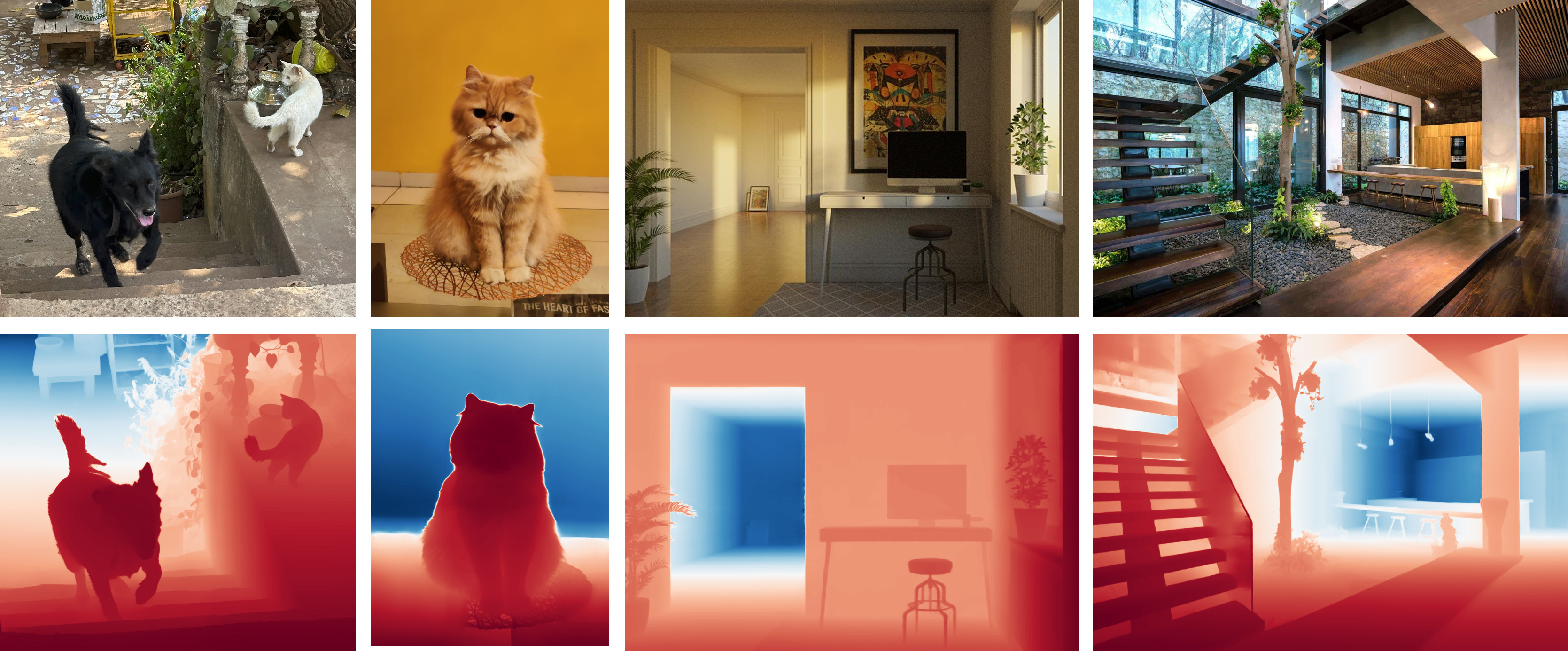}  % Replace with your figure file name
    % \includesvg[width=0.5\textwidth]{MetricGold.svg}
    \caption{We present MetricGold, a diffusion model and associated fine-tuning protocol for monocular metric depth estimation. Its core principle is to leverage the rich visual knowledge stored in modern generative image models. Our model, derived from Stable Diffusion and fine-tuned with photorealistic synthetic data, can zero-shot transfer to unseen datasets, offering sharp monocular metric depth estimation results.}
    \label{fig:your_label}
\end{figure}

\begin{abstract}
Recovering metric depth from a single image remains a fundamental challenge in computer vision, requiring understanding of both scene and scale. While deep learning has advanced monocular relative depth estimation, current models often struggle with unfamiliar scenes and layouts, particularly in zero-shot scenarios and when predicting metric depth. We present MetricGold, a novel approach that harnesses generative diffusion model's rich priors to improve metric depth estimation. Building upon recent advances \cite{marigold, saxena2023zeroshotmetricdepthfieldofview, yang2024depthv2} in MariGold, DMD, and Depth Anything V2 respectively, our method combines latent diffusion, log-scaled metric depth representation, and synthetic data training. MetricGold trains efficient on a single RTX 3090 within two days using photo-realistic synthetic data from \cite{roberts2021hypersim, tartan, cabon2020virtualkitti2}, HyperSIM, VirtualKitti, and TartanAir respectively. Our experiments demonstrate robust generalization across diverse datasets, producing sharper and accurate metric depth estimates compared to existing approaches.
\end{abstract}
\vspace{-1em}
\section{Introduction}

Monocular metric depth estimation seeks to transform a single photographic image into a metric depth map, meaning it regresses a depth value for every pixel. This task is essential whenever understanding the 3D structure of a scene is required, but direct range or stereo measurements are unavailable. However, recovering the 3D structure from a 2D image is a geometrically ill-posed problem, as all the depth information is lost in the projection process. Solving this issue requires prior knowledge, such as understanding of object shapes, sizes, common scene layouts, and occlusion patterns. In other words, monocular metric depth estimation implicitly demands an understanding of both scene geometry and scale. The emergence of deep learning has led to significant improvements in this field. Depth estimation is now treated as a neural image-to-image translation problem, typically learned in a supervised or semi-supervised manner using collections of paired, aligned RGB images and depth maps. Early approaches in this domain were limited by their training data, often focusing on specific environments like indoor scenes \cite{roberts2021hypersim} or driving scenarios \cite{cabon2020virtualkitti2}. 

More recently, there has been a growing interest in developing general-purpose models, such as Metric Depth V2 \cite{yin2023metric3d} for metric depth and Marigold and Depth Anything V1 and V2 \cite{marigold, liu2023vadepth, yang2024depthv2} for relative depth, which generalize to a wide range of scenes without retraining or fine-tuned to specific applications with minimal data. These models often build on the strategy first introduced by MiDaS \cite{Ranftl2020_midas}, which achieves generalization by training high-capacity models on data sourced from diverse RGB-D datasets across multiple domains. The latest advancements include a shift from convolutional encoder-decoder architectures \cite{Ranftl2020_midas} to larger, more powerful vision transformers \cite{ranftl2021_dpt}, and incorporating surrogate tasks \cite{eftekhar2021omnidata} to enhance the model’s understanding of the visual world, leading to better depth predictions. For example, Depth Anything builds on this concept by training a DPT-style transformer for relative depth prediction over DINOV2 embeddings \cite{oquab2024dinov2learningrobustvisual}, while Depth Anything V2 introduces training on photo-realistic simulation datasets, such as \cite{tartan}, \cite{roberts2021hypersim}, and \cite{cabon2020virtualkitti2}, instead of real world datasets. They argue that sensor data introduces biases from sensor-specific corruptions, which the model might learn, potentially leading to suboptimal performance. Their results demonstrate that models trained on photo-realistic datasets produce sharper and more accurate depth predictions. Additionally, visual depth cues depend not only on the scene content but also on the (often unknown) camera intrinsics \cite{yuan2022newcrfs}. As shown in \cite{yang2024depthv2}, relative depth can be predicted with strong generalization and accuracy, pushing the frontier of metric depth estimation even further.

The intuition and implementation behind this work is highly derived from previous these papers \cite{marigold, yang2024depthv2, saxena2023zeroshotmetricdepthfieldofview} the following: Modern image diffusion models have been trained on internet-scale image collections specifically to generate high-quality images across a wide array of domains \cite{betker_dalle3improving_nodate, rombach2022high, saharia2022photorealistic}. If monocular metric depth estimation requires a comprehensive, encyclopedic representation of the visual world, then it should be possible to derive a broadly applicable depth estimator from a pretrained image diffusion model. In this paper, we set out to explore this option and develop MetricGold, a latent diffusion model (LDM) based on Stable Diffusion \cite{rombach2022high}, along with a fine-tuning protocol to adapt the model for metric depth estimation. The key to unlocking the potential of a pretrained diffusion model is to keep its latent space intact. We find this can be done efficiently by modifying and fine-tuning only the denoising U-Net. Turning Stable Diffusion into MetricGold requires only synthetic RGB-D data (in our case, the Hypersim \cite{roberts2021hypersim} and Virtual KITTI \cite{cabon2020virtualkitti2} datasets) and a few GPU days on a single consumer graphics card. Empowered by the underlying diffusion prior of natural images, MetricGold exhibits excellent zero-shot generalization: Without ever having seen real depth maps, it attains generalization on several real datasets.

\section{Re-purposing Image Diffusion Models for Metric Depth}
We present a method that leverages text-to-image diffusion models as an initialization for training a diffusion model to predict metric depth from an image. Specifically, we reformulate monocular metric depth estimation as a generative task, translating RGB images to depth maps using denoising diffusion. Additionally, we demonstrate qualitative results and describe the training process for re-purposing components from the text-to-image diffusion model.

\subsection{Latent Diffusion Models for Depth}
Diffusion models are probabilistic models that assume a forward noising process, progressively transforming a target distribution into a noise distribution. A neural denoiser is then trained to reverse this process iteratively, converting noise samples back into samples from the target distribution. These models have demonstrated exceptional effectiveness with images, videos, and relative depth. They are well-suited for this task because they achieve strong performance on regression problems without the need for specialized architectures, loss functions, or task-specific training procedures. DMD and Marigold \cite{saxena2023zeroshotmetricdepthfieldofview, marigold} have shown how diffusion-based image generators can be fine-tuned for Metric Depth and Affine Invariant Depth with high fidelity generation capabilities. DMD is the closest to our work. DMD does denoising directly in the depth space and we rather choose to do so in the latent space. This makes MetricGold more compute efficient and faster to infer.

\subsection{Using Photo-realistic Synthetic Datasets}
Building on the pioneering work of MiDaS \cite{Ranftl2020_midas} in zero-shot relative depth estimation, recent approaches have focused on creating large-scale training datasets to improve estimation accuracy and generalization. Notably, Depth Anything V1, Metric3D V2 \cite{yin2023metric3d}, and ZeroDepth \cite{guizilini2023towards_zero_depth} have compiled datasets ranging from 1M to 16M images sourced from various sensors. However, the data collected from different sensor suites inherently includes sensor-based noise, depth-related inaccuracies, illumination-induced uncertainty, and depth sensor pattern biases. Although preprocessing and cleaning are applied, these biases persist, acting as hidden contaminants that hinder learning a true depth distribution. Depth Anything V2 \cite{yang2024depthv2} addresses this issue by exclusively training on photo-realistic synthetic datasets, demonstrating sharper and qualitatively superior depth predictions. In our work, we train our model using the Hypersim and Virtual Kitti 2 datasets.

\subsection{Joint Depth scaling for Indoor-Outdoor}
Training a unified model for both indoor and outdoor environments is challenging due to significant variations in depth distribution, color, and scale. Indoor datasets typically contain depths under 10 meters, while outdoor datasets include ground truth depths up to 80 meters. Furthermore, learning a latent space for the diffusion model's metric depth is crucial. To address these issues, we propose using log depth as input to the VAE. Log depth directly addresses the indoor-outdoor depth distribution by applying a scaling function before passing it through the encoder.

\section{Method}
\subsection{Diffusion Formulation}
We frame monocular metric depth estimation as a conditional denoising generation task. MetricGold is trained to model the conditional distribution \( D(d \mid x) \) over depth \( d \in \mathbb{R}^{W \times H} \), where the condition \( x \in \mathbb{R}^{W \times H \times 3} \) is an RGB image.

In the forward process, starting from the initial latent depth map \( d_0 := d \) sampled from the conditional distribution, Gaussian noise is incrementally added at each level \( t \in \{1, \dots, T\} \), yielding progressively noisier samples \( d_t \) defined by:
\[
d_t = \sqrt{\bar{\alpha}_t} d_0 + \sqrt{1 - \bar{\alpha}_t} \, \epsilon
\]
where \( \epsilon \sim \mathcal{N}(0, I) \), \( \bar{\alpha}_t := \prod_{s=1}^{t} (1 - \beta_s) \), and \( \{\beta_1, \dots, \beta_T\} \) represents the variance schedule across the \( T \) diffusion steps. This forward process essentially degrades the initial depth map by adding Gaussian noise at each timestep \( t \), eventually transforming \( d_0 \) into a nearly pure Gaussian noise \( d_T \).

In the reverse process, the conditional denoising model \( \epsilon_{\theta}(\cdot) \), parameterized by learned weights \( \theta \), iteratively removes noise from \( d_t \) to obtain \( d_{t-1} \), gradually reconstructing the original depth structure.

During training, the model parameters \( \theta \) are updated by taking a pair \( (x, d) \) from the training data, adding sampled noise \( \epsilon \) at a randomly chosen timestep \( t \), and computing the noise estimate \( \hat{\epsilon} = \epsilon_{\theta}(d_t, x, t) \). The objective is to minimize one of the denoising diffusion loss functions. The standard noise prediction objective \( L \) is defined as:
\[
L = \mathbb{E}_{d_0, \epsilon \sim \mathcal{N}(0, I), t \sim U(T)} \left[ \| \epsilon - \hat{\epsilon} \|_2^2 \right]
\]
At inference time, the depth map \( d := d_0 \) is reconstructed by starting from a normally distributed variable \( d_T \) and iteratively applying the learned denoiser \( \epsilon_{\theta}(d_t, x, t) \).

% Unlike diffusion models which directly work on the output space, we leverage latent diffusion for multiple factors. Latent Diffusion is compute efficient at training and inference with a trade-off of getting bottelnecked by the reconstruction performance of the depth decoder. The complete diffusion process is retrained as a 

In contrast to traditional diffusion models that operate directly on raw data, latent diffusion models execute diffusion steps within a low-dimensional latent space. This approach enhances computational efficiency and is particularly advantageous for generating high-resolution images. The latent space is formed in the bottleneck layer of a variational autoencoder (VAE) that is trained independently from the denoiser, allowing for effective latent space compression and perceptual alignment with the original data space. To apply our formulation in the latent space, for a given log normalized depth map \( d \), the corresponding latent code is derived from the encoder \( E \) as \( z(d) = E(d) \). Once we have the depth latent code, the decoder \( D \) can reconstruct the depth map, expressed as \( \hat{d} = D(z(d)) \). Similarly, the conditioning image \( x \) is translated into the latent space using \( z(x) = E(x) \). The denoiser is then trained to operate within this latent space, denoted by \( \epsilon_{\theta}(z(d)_t, z(x), t) \). The modified inference procedure introduces an additional step, where the decoder \( D \) reconstructs the depth map \( \hat{d} \) from the estimated clean latent representation \( z(d)_0 \), yielding \( \hat{d} = D(z(d)_0) \).

\subsection{Network Architecture} \label{Network Design}

\begin{figure}[H]
    \centering
    \includegraphics[width=1\textwidth]{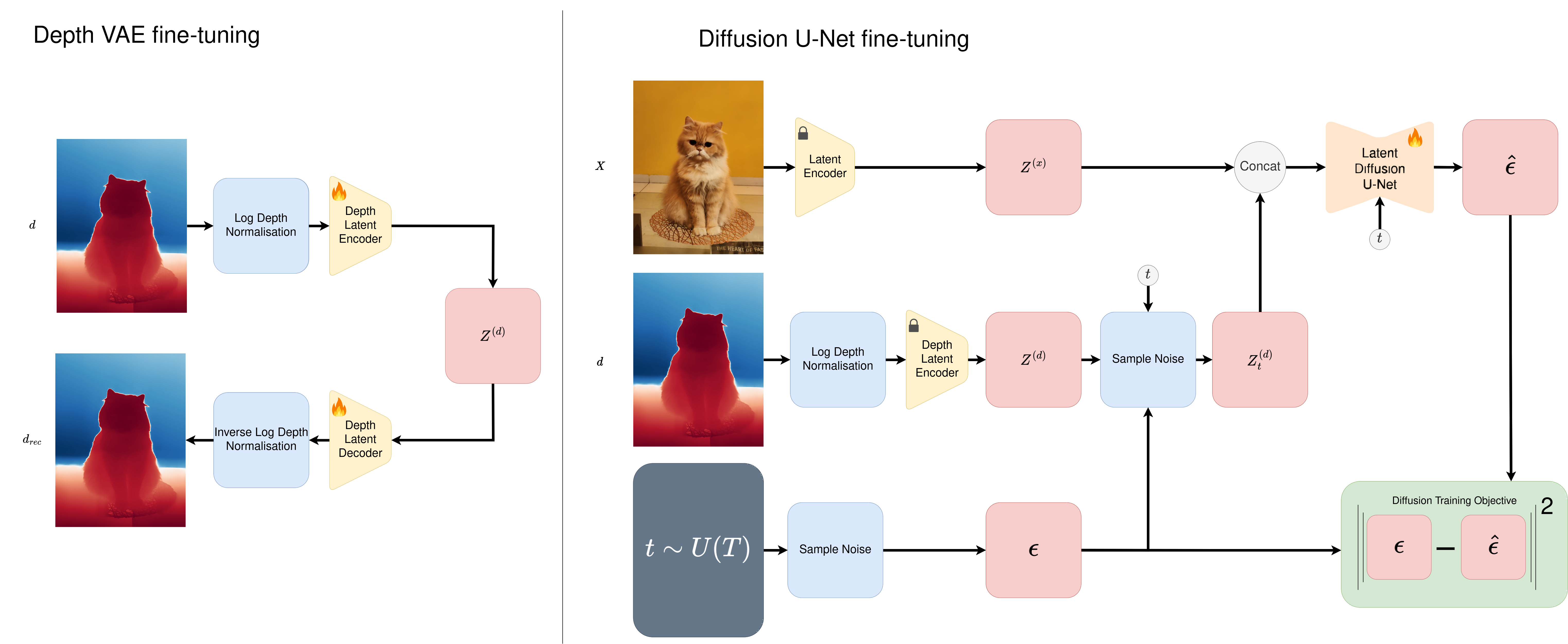}
    \caption{Overview of the MetricGold fine-tuning protocol: Beginning with a pretrained Stable Diffusion model, we first fine-tune the VAE by applying a reconstruction loss on log-normalized metric depth. The image and depth are encoded into their latent spaces using the original Stable Diffusion VAE and the fine-tuned depth VAE, respectively. Next, the U-Net is fine-tuned by optimizing the standard diffusion objective with respect to the depth latent code. To enable image conditioning, the two latent codes are concatenated before being fed into the U-Net, and the first layer of the U-Net is modified to accept the concatenated latent inputs.}
    \label{fig:overview of pipeline}
\end{figure}

One of our main objectives is training efficiency and demonstration that generalizable models can be trained in an academic setting by relying on quality pretrained models for other doamins. WIth minimal changes to the Stable Diffusion v2 model \cite{rombach2022high}, we can add image conditioning. We also finetune Image VAE to learn a Depth VAE to cover the depth distribution in a better manner. \ref{fig:overview of pipeline} contains the overview of the proposed training procedure.

We start by training a Depth VAE, where we initialize our model with the weights of Stable Diffusion's VAE. We log normalize the depth image shown in \eqref{eq:d_linear}.

\paragraph{Log Depth} Most neural networks usually model data distribution in [-1, 1]. One might naively convert metric depth to this range with linear scaling, given we know \(d_{\text{min}}\) and \(d_{\text{max}}\), maximum and minimum range of depth, i.e.,
\begin{equation} \label{eq:normalize}
    normalize = clip(2*d - 1, -1, 1)
\end{equation}
\begin{equation} \label{eq:d_linear}
    d_{\text{linear}} = normalize((d_r - d_{\text{min}})/(d_{\text{max}} - d_{\text{min}})) 
\end{equation}

We find this does not adequately account for the indoor-outdoor depth distribution. To address this, we model the depth using a logarithmic scale, shifting the uncertainty so that lower depth values receive more attention, while larger distances are relatively less important. By using log-scaled depth (\(d_{\text{log}}\)) as the target for inference, we allocate more representation capacity to indoor scenes. 
\begin{equation}
    d_{\text{log}} = normalize(log(d_r/d_{\text{min}})/log(d_{\text{max}}/d_{\text{min}}))
\end{equation}
This adjustment not only improves the representation of depth across both indoor and outdoor distributions, but also leads to enhanced performance of the Variational Autoencoder (VAE). However, we identify that the reconstruction of depth maps from latent embeddings remains the primary performance bottleneck in the pipeline.

\paragraph{Depth encoder and decoder} We utilize the Variational Autoencoder (VAE) from the Stable Diffusion pipeline and fine-tune it specifically for reconstructing log-normalized metric depth. Our analysis has revealed that the reconstruction of the depth image serves as a critical bottleneck in this pipeline, underscoring the importance of this step. Since the model receives a single-channel depth map alongside the 3-channel RGB inputs, we replicate the depth map across the three channels to mimic an RGB image. During inference, the depth latent decoder is employed to generate a three-channel depth image, which is then averaged across the channels to produce the final predicted depth.
\begin{equation}
    \mathcal{L}_{\text{VAE}} = \mathbb{E}_{q(d|x)}\left[ \log p(\log d_{log} |d_{latent}) \right] - D_{KL}(q(d_{latent}|x) \| p(d_{latent}))
\end{equation}

\paragraph{Adapted Denoising U-Net} To implement the conditioning of the latent denoiser \( \epsilon_{\theta}(z(d)_t, z(x), t) \) on the input image \( x \), we concatenate the image and depth latent codes into a single input \( z_t = \text{cat}(z(d)_t, z(x)) \) along the feature dimension. The input channels of the latent denoiser are then doubled to accommodate the expanded input \( z_t \). To prevent inflation of the activation magnitudes in the first layer and to preserve the pre-trained structure as faithfully as possible, we duplicate the weight tensor of the input layer and divide its values by two.

% \subsection{Inference}
% \paragraph{Latent Diffusion Denoising}
% \paragraph{Depth Decoding and De-Normalization}

% \section{Experimentation}
% \subsection{Implementation}
\section{Implementation}
We implement MetricGold using PyTorch and utilize Stable Diffusion v2 as our backbone, adhering to the original pre-training setup with a \( v \)-parameterization objective. We disable text conditioning and follow the steps outlined in \ref{Network Design}. During training, we employ the DDPM noise scheduler with 1000 diffusion steps. At inference time, we switch to the DDIM scheduler and sample only 50 steps. For the final prediction, we aggregate results from 10 inference runs with varying starting noise. Training our method requires 16,000 iterations using a batch size of 32. To accommodate training with a single GPU, we accumulate gradients over 16 steps with a batch size of 2. We use the Adam optimizer with a learning rate of \(5 \times 10^{-5}\). Additionally, we apply random horizontal flipping as an augmentation to the training data. Training our method to convergence takes approximately 2 days on a single Nvidia RTX 3090 GPU.

\section{Conclusion, Further possibilities \& limitations}
We have introduced a fine-tuning pipeline that effectively adapts latent diffusion models for metric depth estimation in a computationally efficient manner. By incorporating incremental innovations such as diffusion for depth, log-normalized depth representation, latent diffusion priors, and synthetic data training, we demonstrate that these strategies can be successfully extended to achieve accurate metric depth predictions. Our approach highlights the versatility of latent diffusion models in solving the challenging task of depth estimation while maintaining efficiency and scalability.

As our pipeline has been trained on \( v \)-parameterization, we can leverage Consistency Distillations \cite{luo2023latentconsistencymodelssynthesizing} methods, to distill the model into lesser number of steps. A paradigm of distilling diffusion model to conditional GANs \cite{kang2024distillingdiffusionmodelsconditional} is also a promising prospect to reduce inference speeds and reduce compute requirements.

% \section{References}
\bibliographystyle{plain}
\bibliography{main}

\end{document}